\documentclass{article}

\usepackage{microtype}
\usepackage{graphicx}
\usepackage{booktabs} % for professional tables

\usepackage[normalem]{ulem}
\usepackage{multirow}
\usepackage{hyperref}

\usepackage[accepted]{icml2024}

\usepackage{amsmath}
\usepackage{amssymb}
\usepackage{mathtools}
\usepackage{amsthm}
\usepackage{subfig}

\usepackage[capitalize,noabbrev]{cleveref}

\theoremstyle{plain}

\theoremstyle{definition}

\theoremstyle{remark}

\usepackage[textsize=tiny]{todonotes}

\icmltitlerunning{}

\begin{document}

\twocolumn[
\icmltitle{How to Alleviate Catastrophic Forgetting in LLMs Finetuning? \\ Hierarchical Layer-Wise and Element-Wise Regularization}



\icmlsetsymbol{equal}{*}

\begin{icmlauthorlist}
\icmlauthor{Shezheng Song}{equal,nudt}
\icmlauthor{Hao Xu }{equal,nudt}
\icmlauthor{Jun Ma}{nudt}
\icmlauthor{Shasha Li}{nudt}
\icmlauthor{Long Peng}{nudt}
\icmlauthor{Qian Wan}{ccnu}
\icmlauthor{Xiaodong Liu}{nudt}
\icmlauthor{Jie Yu}{nudt}
\end{icmlauthorlist}

\icmlaffiliation{nudt}{NUDT}
\icmlaffiliation{ccnu}{CCNU}

\icmlcorrespondingauthor{Jun Ma}{majun@nudt.edu.cn}


\vskip 0.3in
]



\printAffiliationsAndNotice{\icmlEqualContribution} % otherwise use the standard text.

\begin{abstract}
    Large Language Models (LLMs) exhibit strong general language capabilities. However, fine-tuning these models on domain-specific tasks often leads to catastrophic forgetting, where the model overwrites or loses essential knowledge acquired during pretraining. This phenomenon significantly limits the broader applicability of LLMs. To address this challenge, we propose a novel approach to compute the element-wise importance of model parameters crucial for preserving general knowledge during fine-tuning. Our method utilizes a dual-objective optimization strategy: (1) regularization loss based on element-wise parameter importance, which constrains the updates to parameters crucial for general knowledge; (2) cross-entropy loss to adapt to domain-specific tasks. Additionally, we introduce layer-wise coefficients to account for the varying contributions of different layers, dynamically balancing the dual-objective optimization. Extensive experiments on scientific, medical, and physical tasks using GPT-J and LLaMA-3 demonstrate that our approach mitigates catastrophic forgetting while enhancing model adaptability. Compared to previous methods, our solution is approximately 20 times faster and requires only 10\%–15\% of the storage, highlighting the practical efficiency. The code will be released.
\end{abstract}

\section{Introduction}

\begin{figure}
    \centering
    \includegraphics[width=\linewidth]{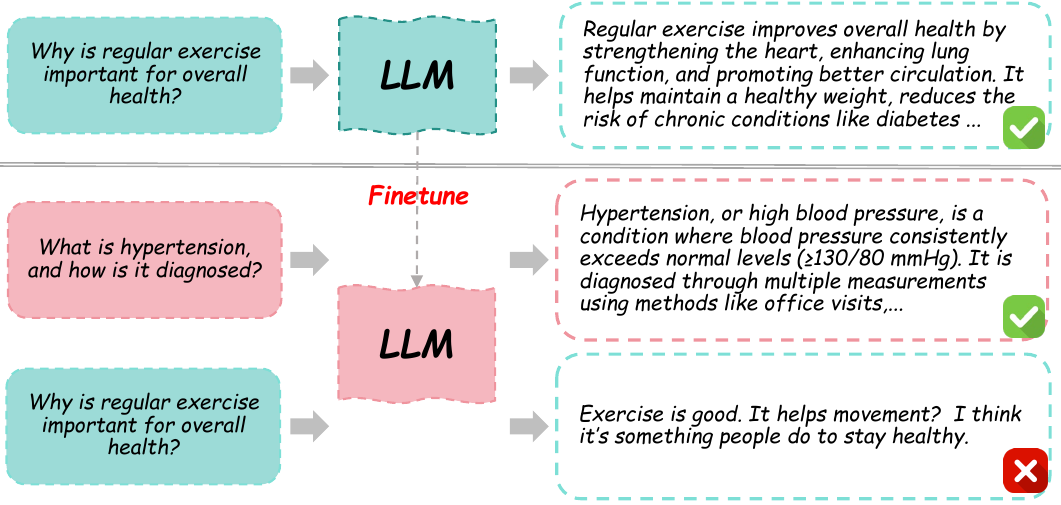}
    \caption{Illustration of catastrophic forgetting: the fine-tuned LLM fails to answer previously known questions.}
    \label{fig:CF_intro}
\end{figure}

Large Language Models (LLMs) are pretrained on massive and diverse datasets, equipping them with remarkable general capabilities~\cite{gpt-j,touvron2023llama2,openai2024gpt4technicalreport}. This pretraining process allows LLMs to serve as versatile tools for a wide range of natural language processing tasks. However, in domains such as medical and scientific fields, LLMs often struggle to perform effectively, necessitating fine-tuning domain-specific data. While fine-tuning could enhance the model task-specific performance, it also introduces a critical challenge: \textbf{catastrophic forgetting}~\cite{Kirkpatrick2016OvercomingCF,kemker2018measuring,Shao20222023,ilora}.
As shown in Figure \ref{fig:CF_intro}, catastrophic forgetting refers to the phenomenon where a model, during the process of fine-tuning, loses or overwrites knowledge learned during pretraining. This issue poses a severe limitation on the broader applicability of LLMs, as it undermines their versatility and reusability across domains.
% 在微调数据中的固定的数据组成和数据格式会损害模型之前所学习到的通用知识，这也会导致模型在完成领域任务时，缺少逻辑推理能力和相关的通用知识，语句上下不连贯。另一方面，会导致在通用任务中，甚至无法回答之前能够回答的问题
The fixed data composition and format in the fine-tuning data may impair the general knowledge previously learned by the model. This results in a loss of logical reasoning abilities and related general knowledge, which affects the model performance on domain-specific tasks. On the other hand, it may also lead to a decline in the ability to answer general tasks, including questions it was previously capable of answering.

% This occurs because the new domain-specific training data typically does not include the diverse and representative information present in the pretraining corpus. Consequently, gradients from the fine-tuning stage disproportionately update parameters critical to the general capabilities of LLMs, leading to a significant degradation in their performance on tasks outside the fine-tuned domain. This issue poses a severe limitation on the broader applicability of LLMs, as it undermines their versatility and reusability across domains.

Addressing catastrophic forgetting is therefore a crucial requirement for maximizing the utility of LLMs. A successful solution needs to achieve a delicate balance: \textbf{retaining the essential general knowledge} when learning new domain-specific expertise. This balance is critical when fine-tuning LLMs for specialized tasks, as both domain adaptation and generalizability are necessary for practical applications.
% TODO EWCLoRA和rsLoRA的做法
% 虽然EWCLoRA关注到了LLM在微调过程中的灾难性遗忘问题，并提出了计算Fisher矩阵以衡量不同参数对于通用能力的重要性。但是其计算所需的时间和空间过于巨大，对于GPTJ-6B来说，计算Fisher矩阵需要22个小时，需要的存储空间为23GB，对于更大参数的LLM来说，所需要的时间和存储空间则更加巨大。另一方面，rsLoRA关注到了模型在继续学习中的稳定性，通过引入一种Rank-Stabilized的缩放因子，使得微调过程中保持了稳定和非崩溃的学习状态，但是其对于通用能力保护的作用并未达到预期效果。
EWCLoRA \cite{ewclora} focuses on the issue of catastrophic forgetting in LLM fine-tuning and uses the Fisher matrix to measure the importance of parameters for general capabilities. 
However, it requires gradients computed with labels from the model distribution, necessitating an additional backpropagation pass for online computation. 
Thus, its computational cost is very high. For GPT-J-6B, calculating the Fisher matrix takes 22 hours on an A800 and requires 23GB of storage, and these requirements increase for larger LLMs. Besides, rsLoRA \cite{rslora} aims to stabilize learning by introducing a rank-stabilized scaling factor, but it does not effectively protect general capabilities as expected.
% Overall, the issue of catastrophic forgetting during fine-tuning still requires attention and improvement.

% Inspired by SI，我们提出一种计算llm内element-wise 参数对于通用能力的影响的方法，保持那些对于通用能力重要的参数，减少对于通用知识的大幅修改。
To address catastrophic forgetting, we calculate parameter importance from two dimensions—element-wise and layer-wise—to constrain the updates of parameters crucial for general capabilities during fine-tuning.
Firstly, our approach calculates the path integral during parameter updates as the \textbf{element-wise parameter importance for regularization} on the general capabilities of the LLM. This helps preserve parameters critical for general knowledge, minimizing significant modifications to it.
Our method could avoid the computation and storage of the Fisher matrix, enabling faster and more storage-efficient computation of parameter importance.
% To address the catastrophic forgetting, inspired by Synaptic Intelligence (SI) \cite{SI}, we propose a method to compute the \textbf{element-wise parameter importance for regularization} on general capabilities of LLM, preserving those parameters that are critical for general knowledge to minimize significant modifications to general knowledge.
Specifically, we define domain $\nu$ as the general knowledge, representing the general capabilities of LLMs, and domain $\mu$ as the knowledge learned during fine-tuning for specific tasks. Our approach leverages a dual-objective optimization strategy that combines two losses: regularization loss, which reduces updates to parameters critical for domain $\nu$ to preserve general knowledge; and cross-entropy (CE) loss, which encourages alignment of domain $\mu$ parameters to enhance domain-specific learning. 
Through the constraint of a dual-objective loss, we aim to maintain general capabilities while performing domain-specific fine-tuning.

Besides, we propose a \textbf{layer-wise coefficient} to adjust the weight of regularization loss.
In LLMs, different layers contribute differently to generalization ability and domain-specific ability. The impact of each layer on the learning process is not uniform; some layers capture high-level abstract features, while others focus on lower-level details. Traditional approaches often treat the importance of each layer as equal, which overlooks the varying degrees of influence that different layers have on the model learning and generalization ability. 
Thus, we propose layer-wise coefficients to dynamically adjust the balance between task learning and the retention of general knowledge in each layer, allowing some layers to prioritize task learning, while others preserve general knowledge. We use the L2 norm of the computed element-wise importance of each layer weight to capture their contribution to both objectives.

Through extensive experiments on scientific, physical, and medical tasks using LLMs (GPT-J and LLaMA-3), we demonstrate that our framework achieves state-of-the-art performance, mitigating catastrophic forgetting while enhancing LLM adaptability.
To maintain general capabilities, it is essential to identify and quantify the importance of various parameters that contribute to these capabilities. The computation of parameter importance is typically time-consuming, and storing the associated weights requires substantial memory resources. Our experimental results demonstrate that our method is nearly \textbf{20 times faster} and requires only \textbf{10\%$\sim$15\% of the storage memory} compared to the previous method, demonstrating the practicality of our approach.
Our contributions are as follows:
\begin{itemize}
    \item We introduce a framework that first records parameter importance on general data, and then applies regularization constraints during fine-tuning on domain-specific data to effectively address catastrophic forgetting in large language models (LLMs).
    \item We propose the element-wise and layer-wise importance metrics to dynamically adjust parameter updates, preserving critical general knowledge while allowing domain-specific expertise to be learned effectively.
    \item Our method achieves state-of-the-art performance across multiple datasets using mainstream backbone LLMs. It significantly reduces computational time (20x faster) and storage (10\%$\sim$15\%) for parameter importance estimation compared to prior methods.
\end{itemize}

\section{Related Work}
\subsection{Continual Learning}
Traditionally, continual learning~\cite{Wickramasinghe20242526,Hadsell20201028,Wickramasinghe20242526,Vijayan2021156} refers to developing learning algorithms to accumulate knowledge on non-stationary data. In general, continuous learning could be categorized into the following methods:
\textbf{Regularization-based methods.} Synaptic Intelligence (SI)~\cite{SI} dynamically estimates the importance of each parameter in an online fashion, penalizing significant changes to parameters that are important for previously learned tasks during training on new tasks.
This method adjusts the learning rate for parameters, ensuring that important parameters are not excessively modified.
Elastic Weight Consolidation (EWC)~\cite{Kirkpatrick2016OvercomingCF} grounded in a Bayesian perspective, estimates the importance of parameters by calculating the Fisher Information Matrix. During new task training, EWC introduces a regularization term that restricts the updates to important parameters, thereby preventing catastrophic forgetting.
From a probabilistic viewpoint, EWC derives an importance matrix that quantifies the significance of network parameters for previous tasks.
\textbf{Architecture-based methods.} Researches
\cite{llamapro,Wang202336193,l2p,Chen20222134} learn new tasks by adapting the structure of existing models. For instance, \citeauthor{l2p} inserts trainable task-specific prompts to the input layer to expand the domain ability. \textbf{Replay-based methods.} Researchers \cite{replay0,replay1,replay2,replay3} retain a subset of previously encountered data, which are reintegrated into the training process of the new tasks. \textbf{Distillation-based methods.} Researches \cite{li2017learning,Cao20213964,Shao20222023,Gu20221707,lfpt5} learn new tasks under the guidance of a teacher model. For instance, Learning without Forgetting (LwF)~\cite{li2017learning} transfers knowledge from old tasks to new tasks, allowing the model to retain performance on the previous task while learning new ones.

\subsection{Catastrophic Forgetting in LLM and LoRA}

With the rapid advancement of large language models (LLMs) \cite{touvron2023llama1, touvron2023llama2}, directly using pretrained models for domain-specific tasks has become prohibitively expensive. As a result, fine-tuning has become the preferred approach, typically divided into full parameter tuning and parameter-efficient fine-tuning (PEFT) methods, such as LoRA (Low-Rank Adaptation) \cite{hu2021lora, wang2024lorapro}.
Full parameter fine-tuning~\cite{lv2023fullpeft} updates all model parameters to improve task adaptability but often causes catastrophic forgetting. PEFT methods like LoRA, by updating only a small subset of parameters through low-rank matrices, reduce computational costs and mitigate forgetting, though some still occur.

To further reduce catastrophic forgetting, researchers have proposed combining EWC with LoRA in a method known as EWCLoRA \cite{ewclora}. This method leverages EWC to calculate the Fisher Information Matrix for parameter importance and uses low-rank matrices of LoRA to limit the scope of parameter updates. However, the calculation of the Fisher matrix introduces significant computational and memory overhead.
Additionally, an interpolation-based LoRA (I-LoRA) method is introduced by \citeauthor{ilora}. I-LoRA constructs a dual-memory experience replay framework, utilizing LoRA parameter interpolation to simulate the weight interpolation process. However, it requires maintaining an additional set of LoRA parameters throughout the process, increasing space cost. 
% Besides, researchers have proposed orthogonal low-rank adaptation (O-LoRA) \cite{olora}. When learning a new task, O-LoRA constrains the LoRA subspace of the new task to be orthogonal to the LoRA subspaces of previous tasks, effectively reducing catastrophic forgetting. 

% 去掉重复图
% \begin{figure}
%     \centering
%     \includegraphics[width=\linewidth]{Figures/loss.png}
%     \caption{Schematic illustration of the loss of cross-entropy (domain) and regularization (general). Total loss is the weighted sum of two losses.}
%     \label{fig:loss_curve}
% \end{figure}

\section{Preliminary}

LoRA is a lightweight and parameter-efficient fine-tuning method that introduces low-rank decomposition into the weight matrix $\theta$ of a pretrained model. Only the newly added low-rank matrices $B$ and $ A $ are optimized, while the main weight $\theta_0$ remains frozen. The parameter at time $t$ during fine-tuning can be expressed as $\theta_t = \theta_0 + \delta_t; \delta_t = B_tA_t$,
% \begin{equation}
%     \theta_t = \theta_0 + \delta_t, \quad \delta_t = B_tA_t
% \end{equation}
where $ \theta_0 \in \mathbb{R}^{d \times d} $ are pretrained weights; $ B \in \mathbb{R}^{d \times r}, A \in \mathbb{R}^{r \times d}$ are the low-rank matrices with $ r \ll d $.

The optimization objective of LoRA is given by:
\begin{equation}
    \mathcal{L}_{\text{LoRA}} = \mathcal{L}(y, f(x; \theta(t)))
\end{equation}
where $ \mathcal{L} $ is the task-specific loss function.

Although LoRA achieves parameter efficiency and training effectiveness, it suffers from catastrophic forgetting, where fine-tuning specific tasks hurts the general ability. 

\begin{figure*}
    \centering
    \includegraphics[width=\linewidth]{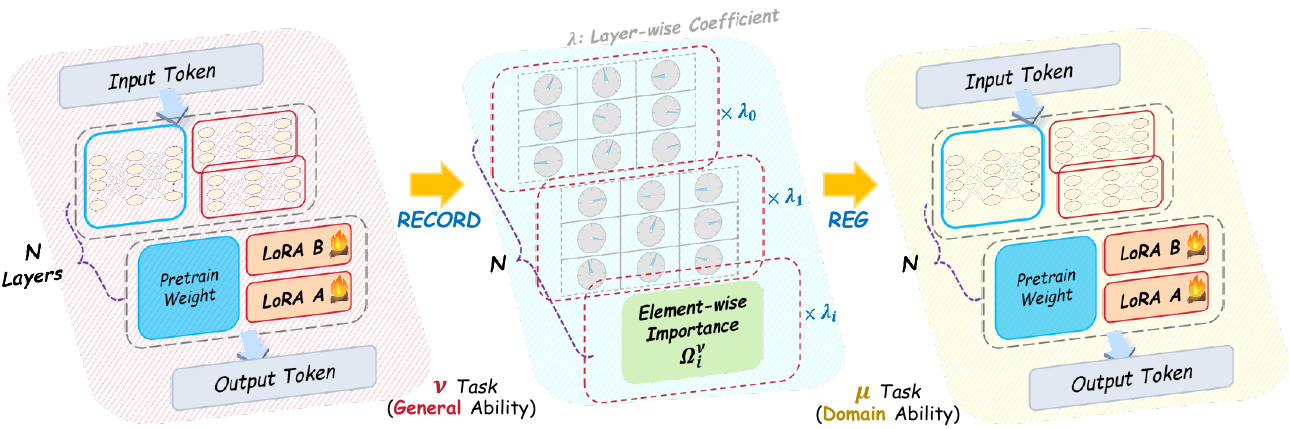}
    \caption{Adaptive constraint combining element-wise and layer-wise importance to preserve general capabilities from the $\nu$ task while learning domain-specific abilities for the $\mu$ task.
    RECORD means the general importance recording in \cref{sec:recording}. REG means the regularization in \cref{sec:element_reg} and \cref{sec:layer_reg}.
    }
    \label{fig:model}
\end{figure*}

\section{Hierarchical Importance Regularization}
% Finetuning with Preserved Generality
Inspired by Synaptic Intelligence (SI) \cite{SI}, we propose a framework to constrain LLMs from making significant changes to their general capabilities during fine-tuning, thus addressing catastrophic forgetting in LoRA tuning.
As shown in Figure \ref{fig:model}, the framework is to compute the importance of each parameter during the training of the initial general task (e.g. $\nu$) and constrain their updates when fine-tuning on subsequent task (e.g. $\mu$). Specifically, the importance scores measure how much each parameter contributes to reducing the loss in the $\nu$ task, and these scores are used to guide the fine-tuning process for the new $\mu$ task. This ensures that the critical parameters for $\nu$ task are modified to a lesser extent when learning $\mu$ task.
% This ensures that essential parameters for the  task are preserved while allowing less critical parameters to adapt to the  task.

\subsection{General Element-Wise Importance Recording}
\label{sec:recording}
In the general task $\nu$, LoRA fine-tuning is performed by minimizing the task-specific loss $ \mathcal{L}_{\text{task}} $. The training process is characterized by a trajectory $ \theta(t) $ in parameter space. The task-specific loss $ \mathcal{L}_{\text{task}} $ is generally computed using cross-entropy loss.
\begin{equation}
    L_{\nu} = \mathcal{L}_{\text{task}}^{\nu}(y_\nu, f(x_\nu; \theta(t))) = - \sum_{i=1}^{N} y_k \log(p_k)
\end{equation}
where $ y_k $ is the true label (target) for the $ i $-th example, $ p_k $ is the predicted probability of the model for the correct label, and $ N $ is the total number of examples.  

We define the contribution of parameter $i$ to the reduction of the loss function as $\omega_i$. The larger the value of $\omega_i$, the more important the parameter $i$ is for maintaining the performance of the task $\nu$.
The change in the loss function from time $t_0$ to time $t_1$ can be defined as the sum of the contributions of all parameters:
\begin{equation}
L(\theta_{t_1}) - L(\theta_{t_0}) =  -\sum_iw_i
\label{eq:w_definition}
\end{equation}
In accordance with the typical behavior of the loss value, which generally decreases, we introduced a negative sign on the right-hand side of \cref{eq:w_definition} to ensure that the value of $\omega_i$ remains positive. 

During the training process of task $\nu$, the total change in the loss function can be obtained by performing a path integral of the gradient of the loss function with respect to the parameters, that is, the path integral from the initial parameter value $\theta_{t_0}$ to the final parameter value $\theta_{t_1}$:
\begin{equation}
L(\theta_{t_1}) - L(\theta_{t_0}) =  \int_{\theta_{t_0}}^{\theta_{t_1}} g(\theta(t))  d\theta(t)
\label{eq:integral}
\end{equation}
where $g$ represents the gradient of the loss function with respect to the parameters. By expanding $d\theta(t)$ in \cref{eq:integral}, we can derive the following expression:
\begin{equation}
    \begin{aligned}
L(\theta_{t_1}) - L(\theta_{t_0}) &=  \int_{t_0}^{t_1} g(\theta(t)) \theta'(t) dt
\\
&=\sum_i\int_{t_0}^{t_1} g(\theta_i(t)) \theta_i'(t)dt
\label{eq:integraldt}
    \end{aligned}
\end{equation}
In accordance with \cref{eq:w_definition} and \cref{eq:integraldt}, it is deduced that the defined quantity $\omega_i$ corresponds precisely to the negative of the path integral of the gradient $g_i$.

\begin{equation}
w_i =  -\int_{t_0}^{t_1} g(\theta_i(t))\theta'_i(t) dt
\label{eq:wk}
\end{equation}

This indicates that we can represent $ \omega_i $ using the product of  $g(\theta_i(t)) = \frac{\partial L}{\partial \theta_i}$ and $\theta_i'(t) = \frac{\partial \theta_i}{\partial t}$ \cite{SI}.

Considering that LoRA utilizes low-rank matrix approximation for fine-tuning, the parameter updates and gradients need to be adjusted accordingly.

The parameters  updating process of low-rank matrices B and A at time $t+1$ are defined as:
\begin{equation}
    \begin{aligned}
    B(t+1) = B(t) - \eta g^B(t) \\
    A(t+1) = A(t) - \eta g^A(t)
    \end{aligned}
\label{eq:ABlearn}
\end{equation}
where $\eta$ is the learning rate, $ g^A(t) $ and $ g^B(t) $ are the gradients of the loss functions with respect to A and B. Based on \cref{eq:ABlearn}, we derive the following expression:
\begin{equation}
    \begin{aligned}
B(t+1)A(t+1) = B(t)A(t) - \eta g^B(t)A(t) \\
- \eta B(t)g^A(t) + \eta^2 g^B(t)g^A(t)
    \end{aligned}
\label{eq:mul}
\end{equation}

According to the definition of LoRA, the parameters at time $t+1$ and time $t$ are respectively defined as:
\begin{equation}
    \begin{aligned}
    \theta(t+1) &= \theta_0 + B(t+1)A(t+1) \\
    \theta(t) &= \theta_0 + B(t)A(t)
    \end{aligned}
\end{equation}
% \begin{equation}
% \theta(t+1) = \theta_0 + B(t+1)A(t+1)
% \end{equation}
% \begin{equation}
% \theta(t) = \theta_0 + B(t)A(t)
% \end{equation}

Based on \cref{eq:mul}, we derive the change of the parameters, which is expressed in terms of $ g^A(t) $ and $ g^B(t) $:
\begin{equation}
    \begin{aligned}
    \Delta \theta &=\theta(t+1)-\theta(t)
    \\
    &=B(t+1)A(t+1) - B(t)A(t)
    \\
    &=-\eta(g^B(t)A(t) + B(t)g^A(t) - \eta g^B(t)g^A(t))
    \end{aligned}
    \label{delta_ceta}
\end{equation}

According to the definition of batch gradient descent, the change in parameters is the negative product of the gradient and the learning rate. If we regard LoRA as a special form of full fine-tuning, we can assume that there exists a gradient $\tilde{g}(t)$  that completes the parameter update process \cite{wang2024lorapro}.

Based on \cref{delta_ceta} and the definition of $\tilde{g}(t)$, we obtain the  parameter change and hypothetical gradient at time t.
\begin{equation}
\begin{aligned}
\tilde\theta'(t) &= B(t+1)A(t+1) - B(t)A(t) \\
\tilde{g}(t) &=g^B(t)A(t) + B(t)g^A(t) - \eta g^B(t)g^A(t)
\end{aligned}
\end{equation}

% Based on \citeauthor{lorapro}, the equivalent gradient is defined as:
% \begin{equation}
% \tilde{g}_i(t) = B_i(t) g^A_i(t) + g^B_i(t) A_i(t),
% \end{equation}
% where $ g^A $ and $ g^B $ are the gradients with respect to $ A $ and $ B $, respectively.
% Similarly, The change in the parameters $\theta_i'(t)$ can be expressed as:

% \begin{equation}
%     \begin{aligned}
%         \theta_i'(t) &= \theta_i(t) - \theta_i(t-1) = \delta_i(t) - \delta_i(t-1) \\
%     &= B_i(t)A_i(t)-B_i(t-1)A_i(t-1).
%     \end{aligned}
% \end{equation}

% \begin{equation}
% \omega_i^\nu \approx \tilde{g}_i(t) \cdot \theta_i'(t)
% \end{equation}

In this way, we obtain the value of $\omega_i$ for the LoRA scenario. 
\begin{equation}
\begin{aligned}
w_i =  -\int_{t_0}^{t_1} \tilde g_i(t)\tilde \theta'_i(t) dt
\end{aligned}
\end{equation}

To quantify the importance of each parameter, we calculate an importance score $ \Omega_i^\nu $ based on its contribution to the change in loss during training of task $\nu$. Specifically, the importance of a parameter is computed as:
\begin{equation}
\Omega_i^\nu = \sum_{\nu} \frac{\omega_i^\nu}{(\Delta_i^\nu)^2 + \xi}
\label{eq:importance}
\end{equation}
where $ \Delta_i^\nu = \theta_i(t^\nu) - \theta_i(t^0) $ is whole change of the $ i $-th parameter $ \theta_i $ during task $ \nu $, $ \theta_i(t^\nu)$ is the final parameter after fine-tuning on task $\nu$. In the context of LoRA fine-tuning, the  $ \Delta_i^\nu$ is defined as $ (B(t^\nu)A(t^\nu))_i $. This relationship stems from the fact that, at the initialization of LoRA at time $0$, the B matrix is set to zero.
The term in the denominator $ (\Delta_i^\nu)^2 $ ensures that the regularization term carries the same units as the loss $ L $. $ \xi $ is a small positive constant to prevent division by zero. 
This formulation assigns higher scores to parameters that have a significant impact on loss reduction while accounting for their magnitude to avoid bias toward large updates.

\subsection{Element-Wise Regularization in Domain Tuning}
\label{sec:element_reg}

After fine-tuning the $\nu$ task, we extend the optimization objective to include both task-specific and regularization losses during $\mu$ finetuning. The task-specific loss $ \mathcal{L}_{\text{task}}^\mu $ drives the adaptation to the $\mu$ task.
To preserve knowledge from the $\nu$ task, the regularization loss penalizes deviations from the important parameter values recorded in the $\nu$ task. The regularization loss $\mathcal{L}_{\text{reg}, l}^\nu$ of the $l$-th layer is defined as:
\begin{equation}
\mathcal{L}_{\text{reg}, l}^\nu =  \sum_{i}^n  \sum_{\nu<t<\mu} \Omega_i^\nu ( \theta_i^t - \theta_i^\nu )^2
\label{eq:reg_loss_cal}
\end{equation}
where $ \Omega_i^\nu $ represents the importance of the $ i $-th parameter in the $\nu$ task, and $ \theta_i^\nu $ is the reference parameter after $\nu$ task fine-tuning. 
This loss ensures that parameters with high importance scores remain close to their $\nu$ task values while allowing less important parameters more flexibility for adaptation.
During training, $ \omega_i $ values are updated continuously, while the cumulative importance $ \Omega_i^\nu $ is updated only at the end of task $\nu$. After updating $ \Omega_i^\nu $, the $ \omega_i $ is reset to zero.

\subsection{Layer-Wise Coefficient Regularization}
\label{sec:layer_reg}
We compute the importance of each layer based on its contribution to the parameters learned in the $\nu$ task. This layer-specific importance metric allows the model to dynamically adjust the regularization across different layers.
The layer-wise weighted regularization is defined as :
\begin{equation}
   \mathcal{L}_{\text{reg}}^\nu =  \sum_l \text{softmax} (\|\mathbf{\Omega}_l^\nu\|_2) \mathcal{L}_{\text{reg}, l}^\nu
\end{equation}
where $\|\mathbf{\Omega}_l^\nu\|_2$ denotes the L2 norm of the parameter importance matrix $\mathbf{\Omega}_l^\nu$ for the $l$-th layer, which reflects the significance of the parameters learned in the $\nu$ task. 
The total loss for the $\mu$ task is defined as:
\begin{equation}
\mathcal{L}^\mu = \mathcal{L}_{\text{task}}^\mu + \varphi \mathcal{L}_{\text{reg}}^\nu
\label{eq:total_loss}
\end{equation}
The use of this adaptive regularization $\mathcal{L}_{\text{reg}}^\nu$ helps mitigate catastrophic forgetting by maintaining the integrity of essential features learned in prior tasks. $\varphi$ is the hyperparameter that controls the weight of the domain ($\mathcal{L}_{\text{task}}$) and general ($\mathcal{L}_{\text{reg}}$) ability of LLM.

\section{Experiments}

% % Table generated by Excel2LaTeX from sheet 'Sheet1'
\begin{table*}[htb]
  \centering
  \caption{General and domain ability of LLMs.
  (Acc$\uparrow$: Accuracy of domain ability, PPL$\downarrow$: Perplexity of general ability.)
  }
    \resizebox{.95\linewidth}{!}{
    \begin{tabular}{lcccccccccccc}
    \toprule
          & \multicolumn{6}{c}{\textbf{LLaMA-3}}         & \multicolumn{6}{c}{\textbf{GPT-J}} \\
    \midrule
          & \multicolumn{2}{c}{\textbf{SciQ}} & \multicolumn{2}{c}{\textbf{PiQA}} & \multicolumn{2}{c}{\textbf{MedMCQA}} & \multicolumn{2}{c}{\textbf{SciQ}} & \multicolumn{2}{c}{\textbf{PiQA}} & \multicolumn{2}{c}{\textbf{MedMCQA}} \\
          & PPL↓  & Acc↑  & PPL↓  & Acc↑  & PPL↓  & Acc↑  & PPL↓  & Acc↑  & PPL↓  & Acc↑  & PPL↓  & Acc↑ \\
    \midrule
    Base  & 4.94  & 95.10  & 4.94  & 48.53  & 4.94  & 18.50  & 3.28  & 91.60  & 3.28  & 49.13  & 3.28  & 21.30  \\
    \midrule
    LoRA($\mu$)                               & 5.05            & 96.20           & 5.43            & \uline{48.75}     & 5.04             & 53.69             & 3.43      & \uline{96.50}     & 3.54            & \uline{50.16}     & 3.49             & \textbf{38.35}    \\
    LoRA($\nu+\mu$)                             & 5.31            & 96.10           & 5.58            & 46.91           & 5.15             & 53.12             & 3.39            & 96.20           & 3.52            & 49.51           & 3.37             & 33.66    \\
    rsLoRA                                & 5.28            & \uline{96.50}     & 5.71            & 47.50           & 5.24             & 51.92             & 3.50            & 96.20           & 3.65            & 49.62           & \uline{3.35}       & 35.69             \\
    EWC-L                                 & \uline{4.88}      & 96.30           & \uline{4.98}      & 48.45           & \uline{4.79}       & \textbf{56.39}    & \uline{3.38}            & 96.10           & \uline{3.47}      & 49.40           & 3.38             & \uline{36.48}             \\
    \textbf{Ours}                         & \textbf{4.64}   & \textbf{97.10}  & \textbf{4.90}   & \textbf{51.14}  & \textbf{4.64}    & \uline{55.80}       & \textbf{3.35}   & \textbf{96.80}  & \textbf{3.40}   & \textbf{50.49}  & \textbf{3.34}    & 36.10  \\
    \bottomrule
    \end{tabular}%
    }
\label{tab:mainresult}%
\end{table*}%

\subsection{Backbone LLMs and Baseline Methods}
Following the previous work \cite{ewclora}, two mainstream LLMs
are used for the evaluation of our method:
(1) \textit{GPT-J} \cite{gpt-j} is a GPT-2-like causal language model trained on the Pile dataset. It is suitable for various understanding and generation tasks.
(2) \textit{LLaMA-3} \cite{dubey2024llama3} is the third-generation open-source LLM. It is designed with enhanced efficiency and scalability, offering state-of-the-art performance across various benchmarks. 
% \item MiniCPM3 is a lightweight Chinese language model designed for efficiency and adaptability. It leverages pretraining on extensive multilingual corpora and is tailored for tasks requiring domain adaptation. 
These models vary in architecture and parameter count, enabling a robust evaluation of our method.

We compare our method with the following approaches:
(1) \textit{Base}: the model without any tuning.
(2) \textit{LoRA($\mu$)} \cite{hu2021lora}: the method is fine-tuned using only data from the $\mu$ task (domain-specfic task).
(3) \textit{LoRA($\nu + \mu$)}: the method is first fine-tuned using data from the $\nu$ task (general task), and then fine-tuned using data from the $\mu$ task (domain-specific task).
(4) \textit{EWCLoRA} \cite{ewclora}: a method using the EWC method, where the Fisher matrix is computed and regularization constraints are applied to preserve the important parameters while updating for the new task. 
(5) \textit{rsLoRA}: an enhanced LoRA method that modifies the scaling factor to prevent gradient collapse, enabling better fine-tuning performance with higher-rank adapters while maintaining the same inference cost.
% (5) \textit{SLoRA}: our proposed method, leverages a dual-objective optimization strategy to preserve critical parameters for general tasks (via Regularization Loss) and adapt to domain-specific tasks (via Cross-Entropy Loss), ensuring effective learning while mitigating catastrophic forgetting. 

\begin{figure}[htb]
    \centering
    % \subfloat[General Ability]{
    %     \includegraphics[width=.245\textwidth]{Figures/reg_conf/PPL.pdf}
    %     \label{fig:reg_conf_ppl}
    %     }
    \subfloat[SciQ PPL.]{\includegraphics[width=.45\linewidth]{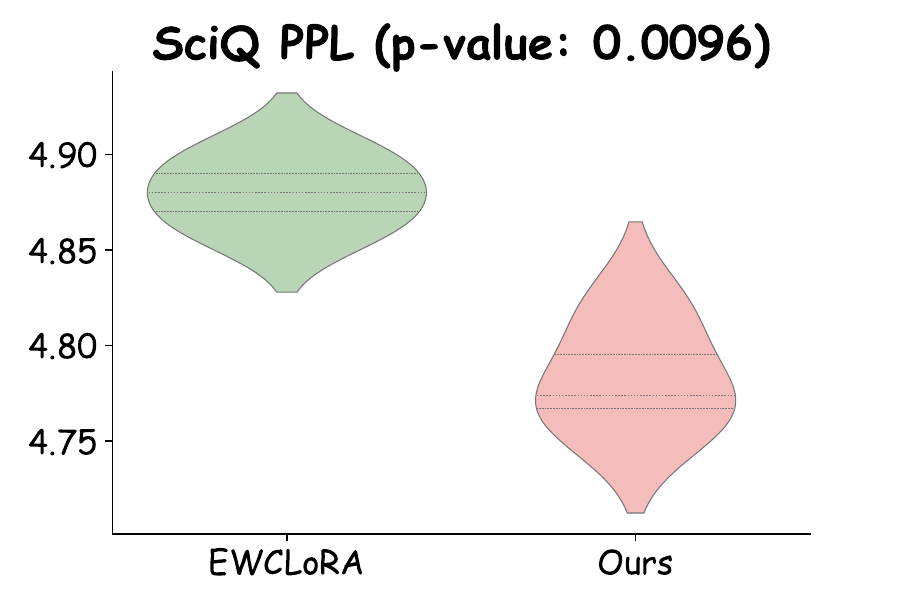}\label{fig:t_test_ppl_sciq}}
    \subfloat[SciQ Acc.]{\includegraphics[width=.45\linewidth]{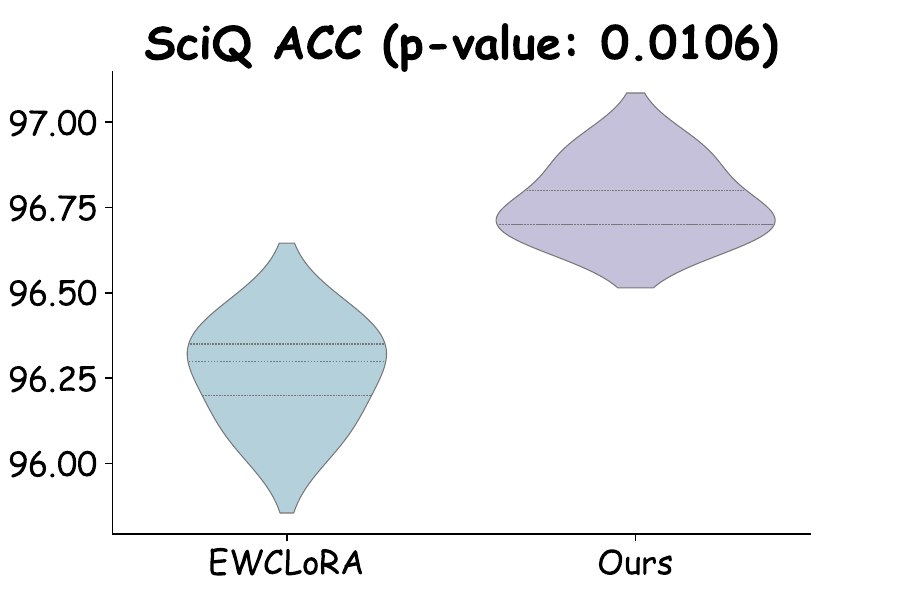}\label{fig:t_test_acc_sciq}}
    \\
    \subfloat[PiQA PPL.]{\includegraphics[width=.45\linewidth]{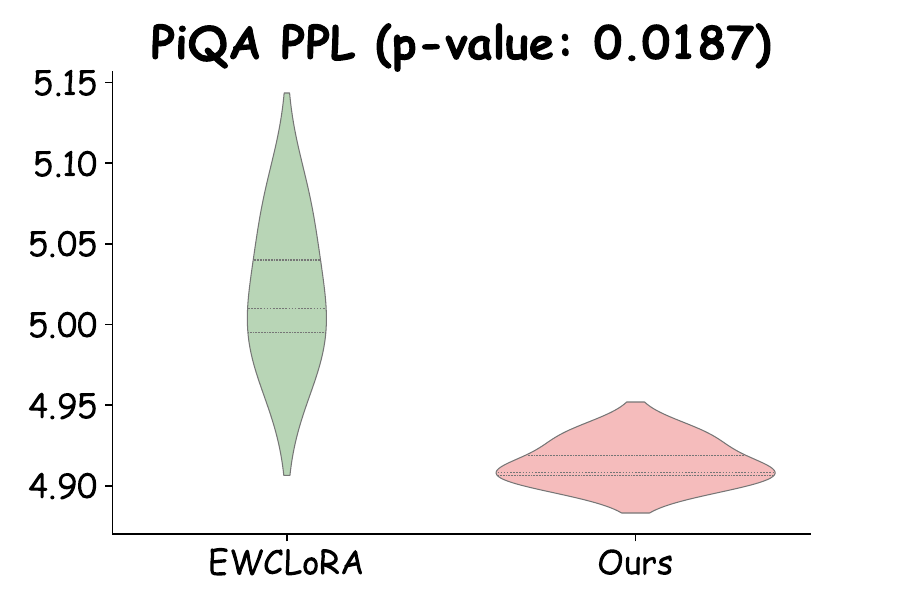}\label{fig:t_test_ppl_piqa}}
    \subfloat[PiQA Acc.]{\includegraphics[width=.45\linewidth]{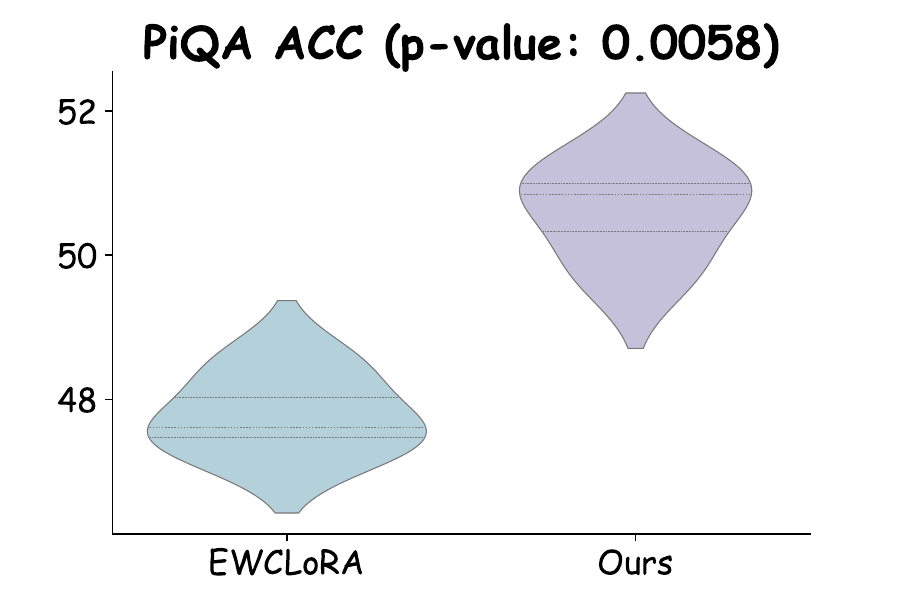}\label{fig:t_test_acc_piqa}}
    \\
    \subfloat[MedMCQA PPL.]{\includegraphics[width=.45\linewidth]{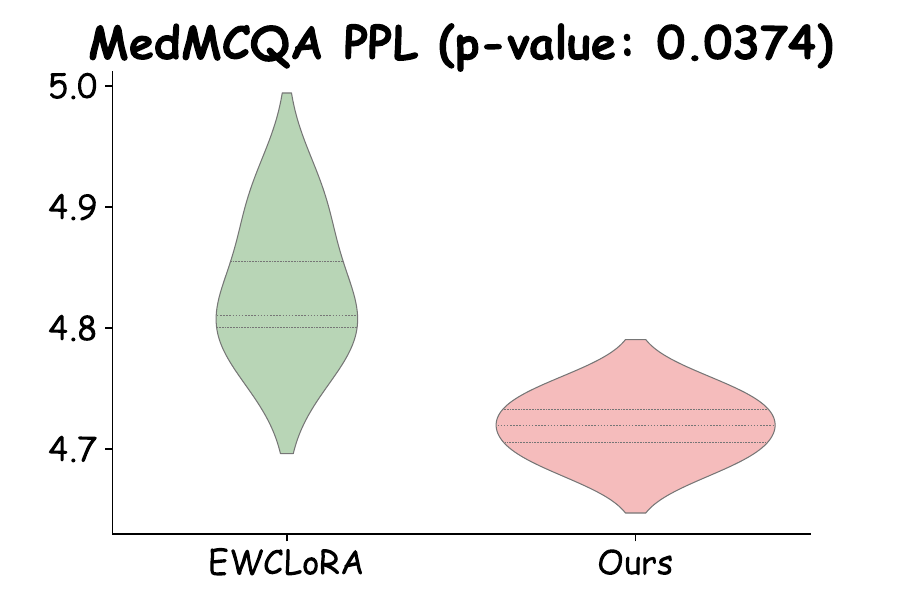}\label{fig:t_test_ppl_medmcqa}}
    \subfloat[MedMCQA Acc.]{\includegraphics[width=.45\linewidth]{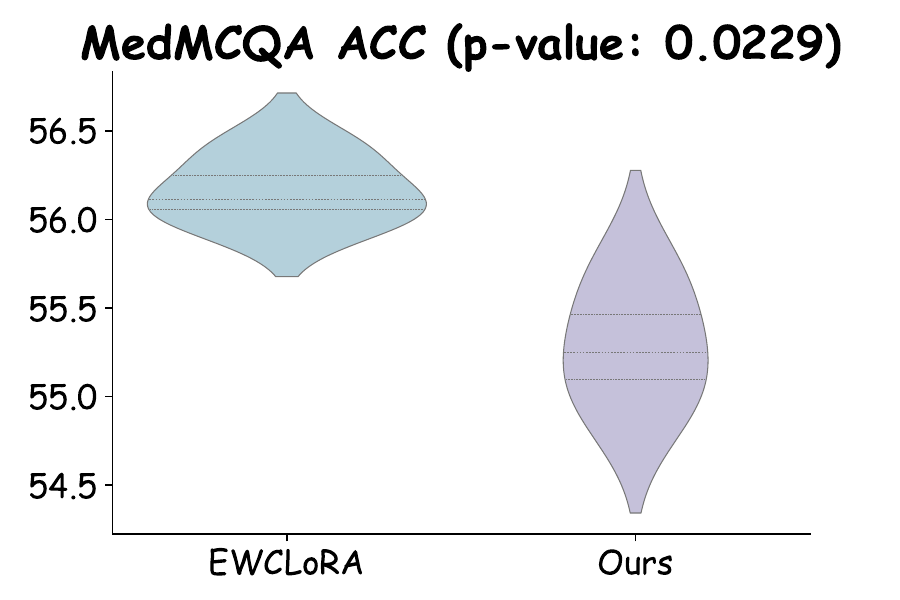}\label{fig:t_test_acc_medmcqa}}
    \caption{Independent samples t-test of EWCLoRA and our method on LLaMA-3: violin plots of perplexity (PPL) and accuracy (Acc) across datasets}
    \label{fig:t_test}
\end{figure}

\subsection{Tasks, Metrics, and Hyperparameters}
\textbf{$\nu$ Task (General Ability):}
The $\nu$ task focuses on learning which parameters are important for general tasks. 
Following previous work \cite{ewclora}, we take Pile \cite{gao2020pile} as the evaluation datasets for LLM general ability. LoRA is applied to fine-tune the model on the $\nu$ task, and parameter importance for Synaptic Intelligence (SI) is recorded during this stage.

\textbf{$\mu$ Task (Domain Ability):}
The $\mu$ task evaluates the ability to adapt to specific tasks while mitigating catastrophic forgetting of general knowledge. We select three representative tasks:
(1) \textit{Medical task:} MedMCQA dataset \cite{pal2022medmcqa}. 
(2) \textit{Scientific task:} SciQ dataset \cite{welbl2017sciq}.
(3) \textit{Physics task:} PiQA dataset \cite{bisk2020piqa}.

% \subsection{Hyperparameters}
The LLMs selected for our experiments are GPT-J-6B and LLaMA 3.2-3B. The batch size is set to 20, and the learning rate is set to 8e-4. The rank for LoRA fine-tuning is set to 8, with the LoRA alpha value set to 32. Both the $\nu$ and $\mu$ tasks are trained for 5 epochs.

\section{Results and Analysis}
\subsection{Comparison of General and Domain Capabilities}
As shown in \cref{tab:mainresult}, our method achieves better preservation of general ability (as reflected by the lowest PPL) while maintaining domain-specific accuracy comparable to, or even better than, previous methods. This demonstrates that our approach effectively balances domain accuracy and general perplexity.

\cref{fig:t_test} presents a comparison between the results of EWCLoRA and our method through independent samples t-tests. The six subplots show the Perplexity (PPL) and Accuracy (Acc) across SciQ, PiQA, and MedMCQA datasets. The p-values for perplexity on SciQ, PiQA, and MedMCQA, and for accuracy on SciQ and PiQA are below 0.05, indicating statistically significant differences and demonstrating the superiority of our method over EWCLoRA.

\begin{figure*}[htb]
    \centering
    \subfloat[LLaMA-3 on SciQ]{\includegraphics[width=.31\textwidth]{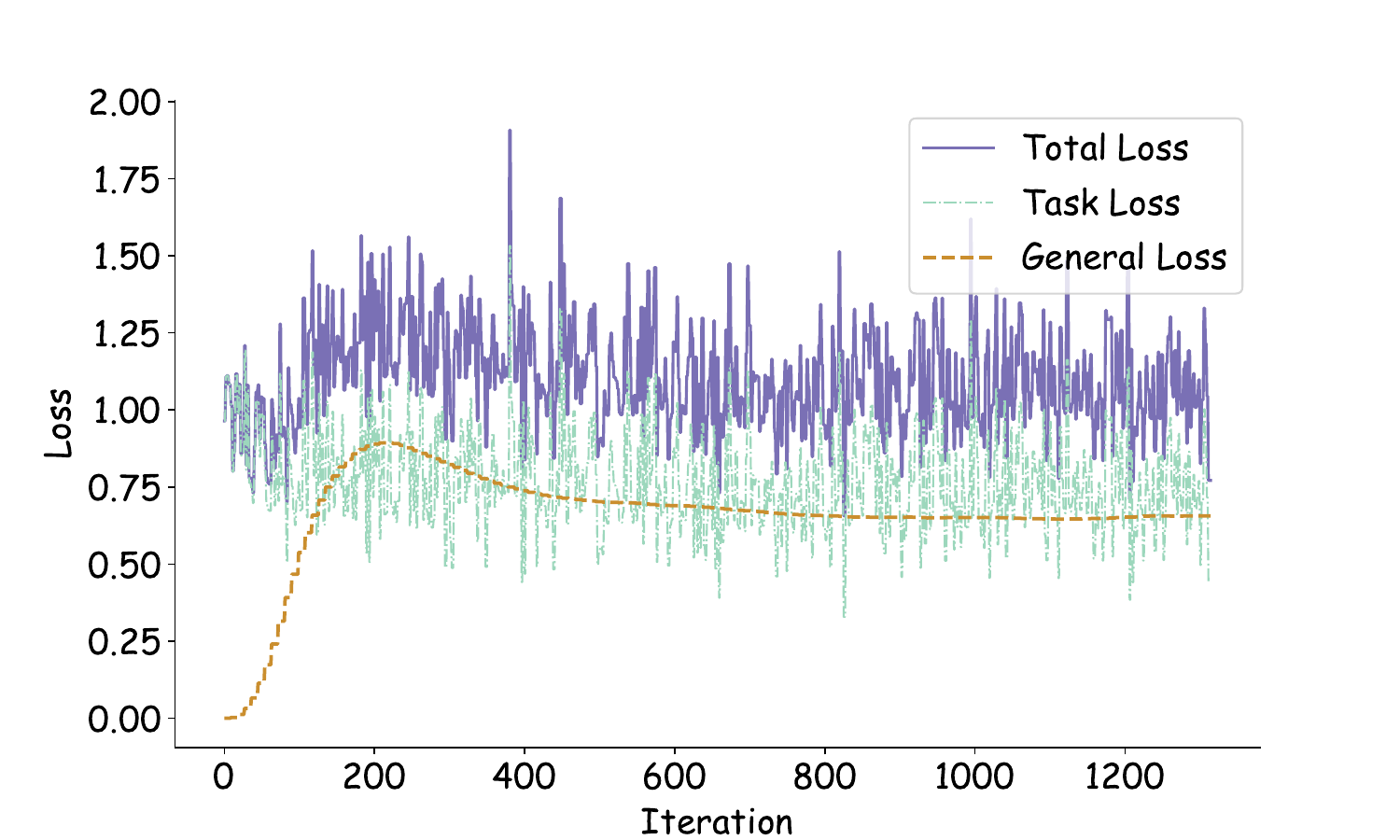}\label{fig:loss_curve_llama_sciq}}
    \subfloat[LLaMA-3 on PiQA]{\includegraphics[width=.31\textwidth]{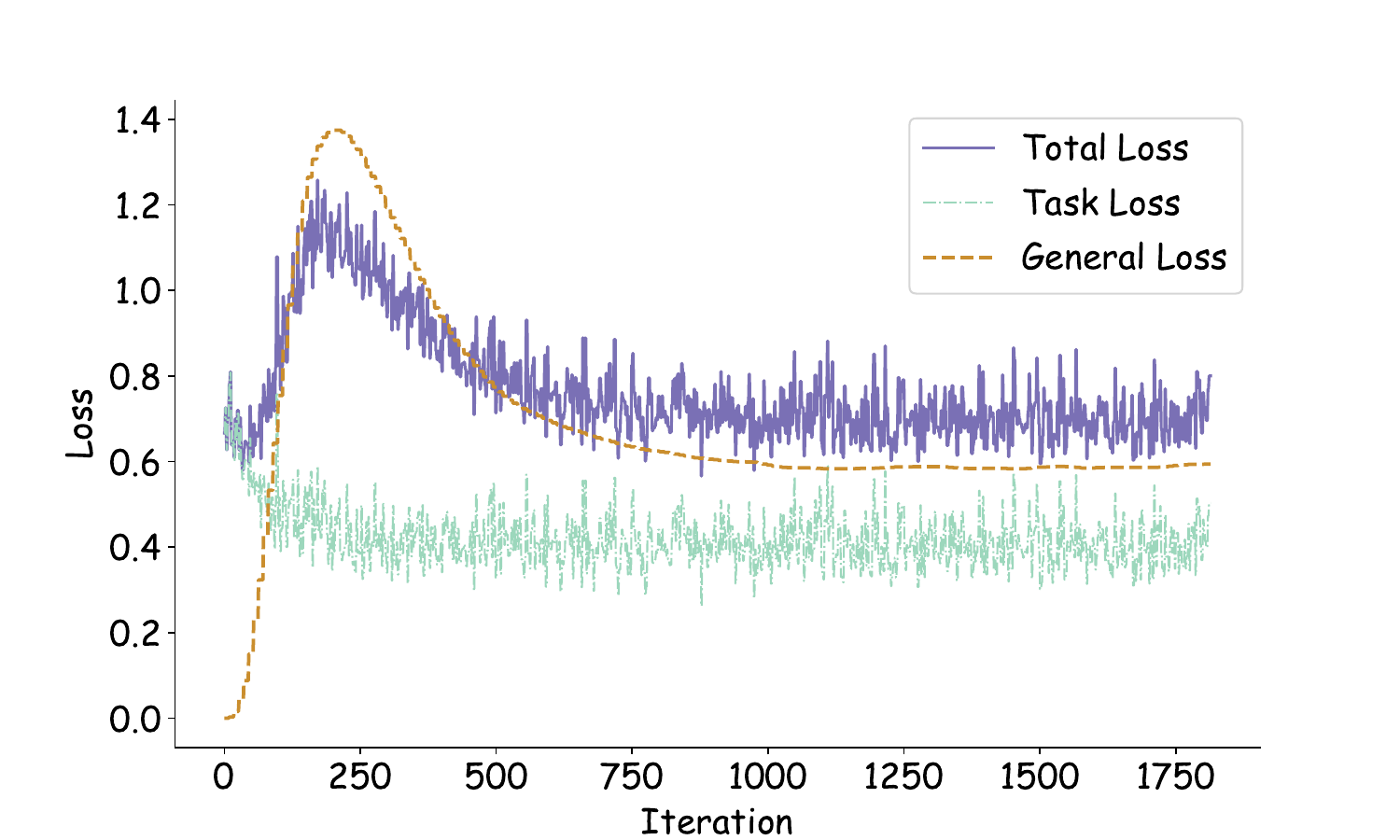}\label{fig:loss_curve_llama_piqa}}
    \subfloat[LLaMA-3 on MedMCQA]{\includegraphics[width=.31\textwidth]{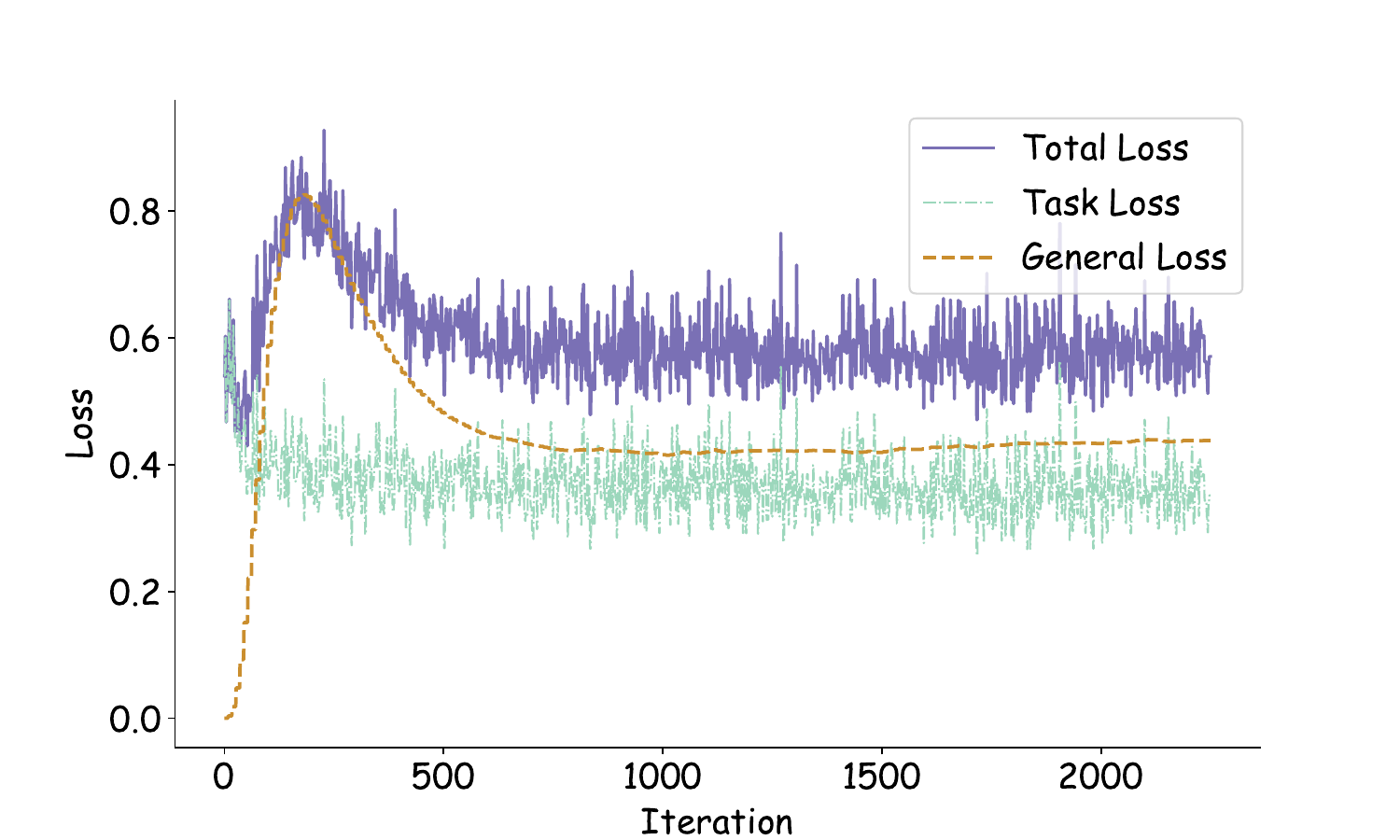}\label{fig:loss_curve_llama_medmcqa}}
    \\
    \subfloat[GPT-J on SciQ]{\includegraphics[width=.31\textwidth]{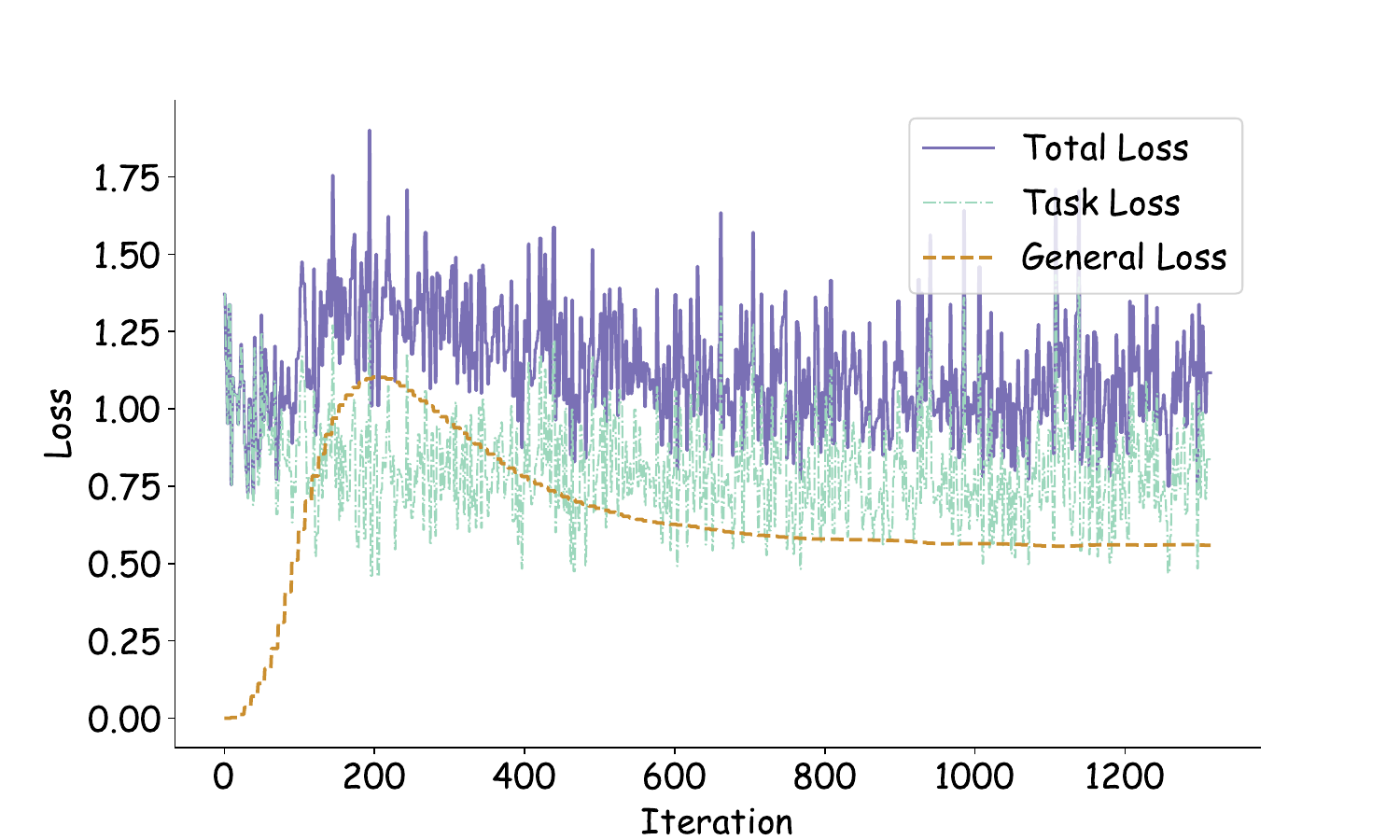}\label{fig:loss_curve_gptj_sciq}}
    \subfloat[GPT-J on PiQA]{\includegraphics[width=.31\textwidth]{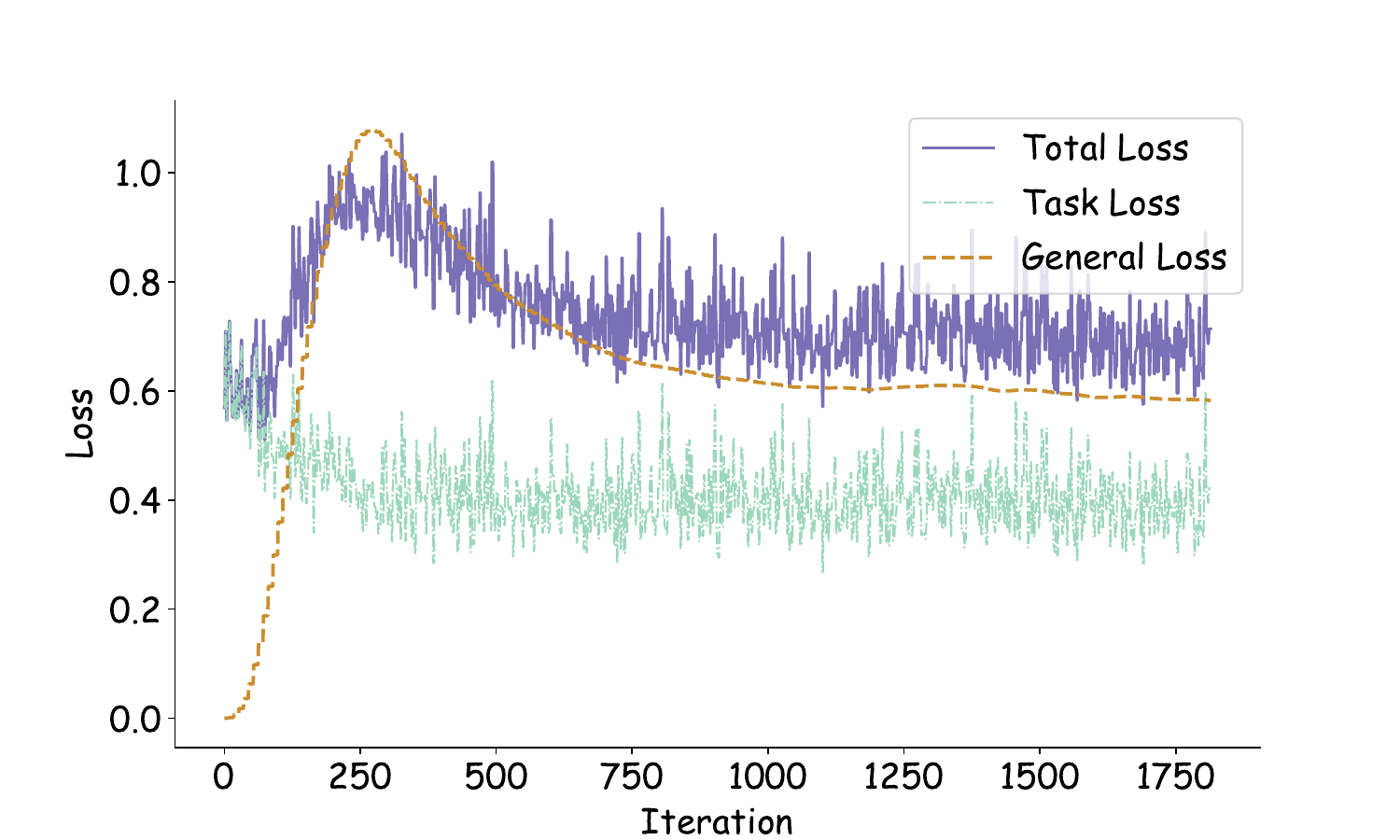}\label{fig:loss_curve_gptj_piqa}}
    \subfloat[GPT-J on MedMCQA]{\includegraphics[width=.31\textwidth]{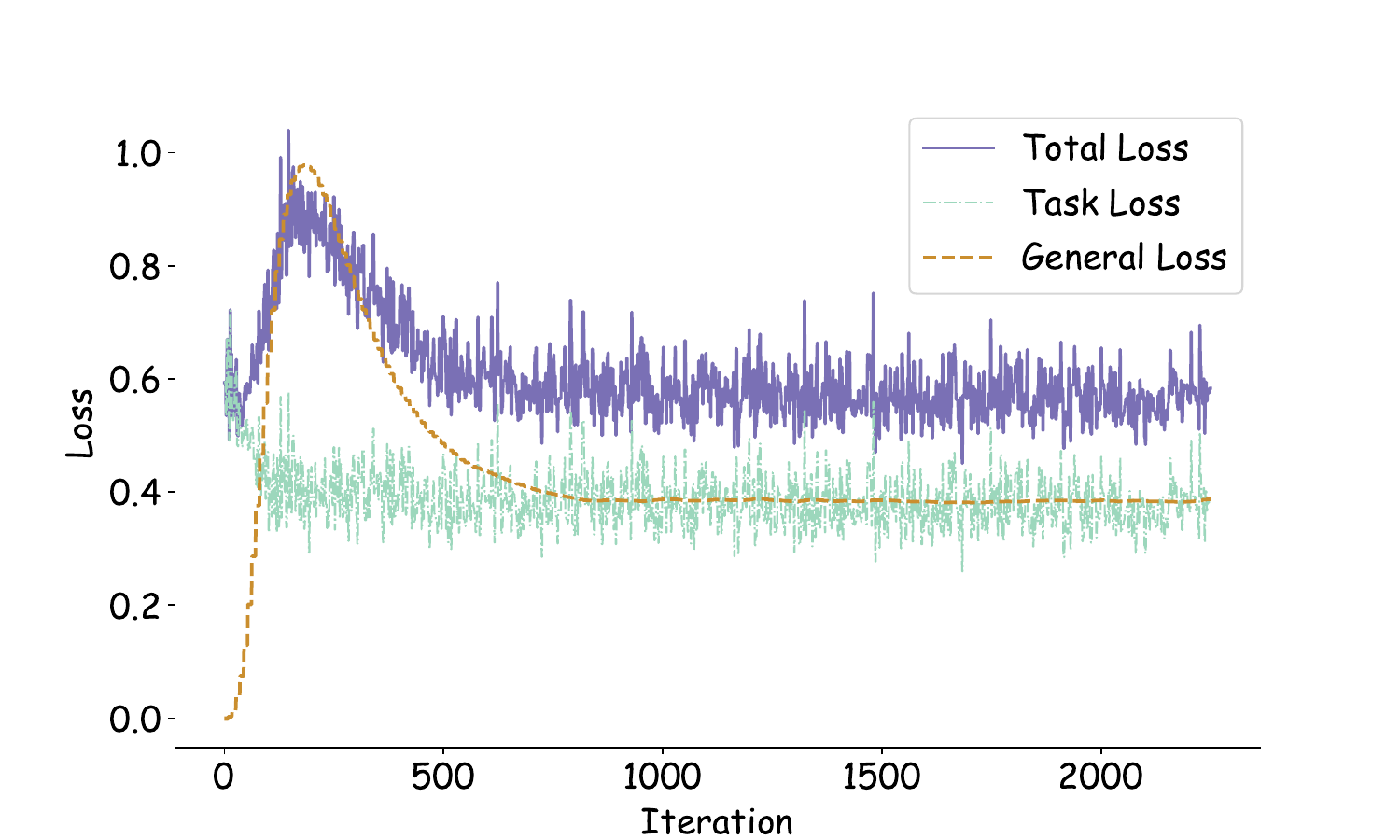}\label{fig:loss_curve_gptj_medmcqa}}
    \caption{Loss curves on three datasets: balancing task learning and generalization. The total loss consists of task loss (\(\mathcal{L}_{\text{task}}\)) and a scaled version of general loss (\(\mathcal{L}_{\text{reg}}\)), where task loss controls the model learning on new domain data, and general loss helps maintain the model generalization ability.}
    \label{fig:loss_curve}
\end{figure*}

% \subsection{Loss Curve Analysis}
\cref{fig:loss_curve} shows the loss curves in the learning process of GPT-J and LLaMA-3 across three datasets. The total loss is the weighted sum of the task loss $\mathcal{L}_{\text{task}}$ and general loss $\mathcal{L}_{\text{reg}}$. As observed, the task loss continuously decreases, while the $\mathcal{L}_{\text{reg}}$ exhibits an initial increase followed by a decrease. 
As defined  in \cref{eq:reg_loss_cal}, $\mathcal{L}_{\text{reg}}$ measures the difference between the model parameters $\theta_{\nu}$ after learning on task $\nu$ and the model parameters $\theta_{\mu}$ learned on the current task $\mu$. Initially, when learning on task $\mu$, the model parameters are not yet updated, so the general loss is zero. As the task loss updates the parameters, the model starts to deviate from $\theta_{\nu}$, causing the general loss to rise. This mechanism enforces the model to learn in a way that minimizes both general and task losses simultaneously.

\begin{figure}[htbp]
    \centering
    \includegraphics[width=.8\linewidth]{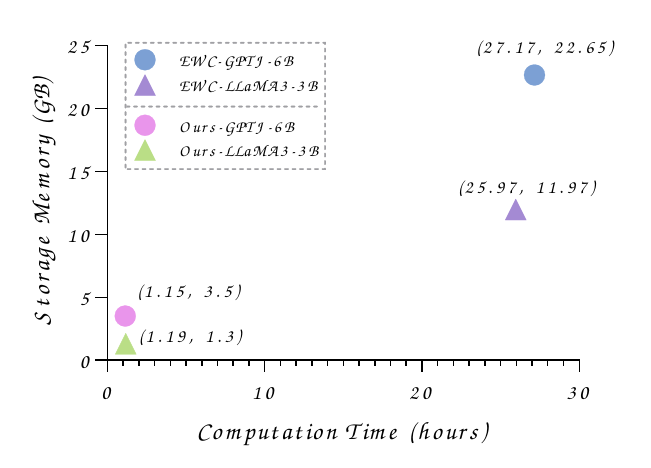}
    \caption{Comparison of computation time and storage for importance $\mathbf{\Omega}_l^\nu$ between previous method and ours.}
    \label{fig:Time Complexity}
\end{figure}

\subsection{Complexity Comparision}

% \begin{figure}[htb]
%         \centering
%         \subfloat[Effect of batch size]{\includegraphics[width=.5\linewidth]{figure/batchszie.pdf}\label{fig:bs}}
%         \subfloat[Effect of loss weight $\beta$]{\includegraphics[width=.5\linewidth]{figure/loss.pdf}\label{fig:loss}}
%         \caption{The statistics and experimental result under different hyperparameters on Twitter15. }
%     \end{figure}

    \begin{figure*}[htb]
        \centering
        % \subfloat[General Ability]{
        %     \includegraphics[width=.245\textwidth]{Figures/reg_conf/PPL.pdf}
        %     \label{fig:reg_conf_ppl}
        %     }
        \subfloat[SciQ]{\includegraphics[width=.33\textwidth]{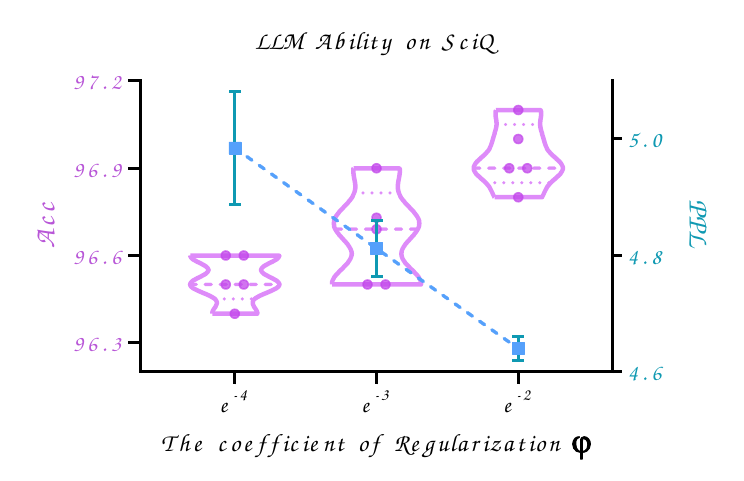}\label{fig:reg_conf_sciq}}
        \subfloat[PiQA]{\includegraphics[width=.33\textwidth]{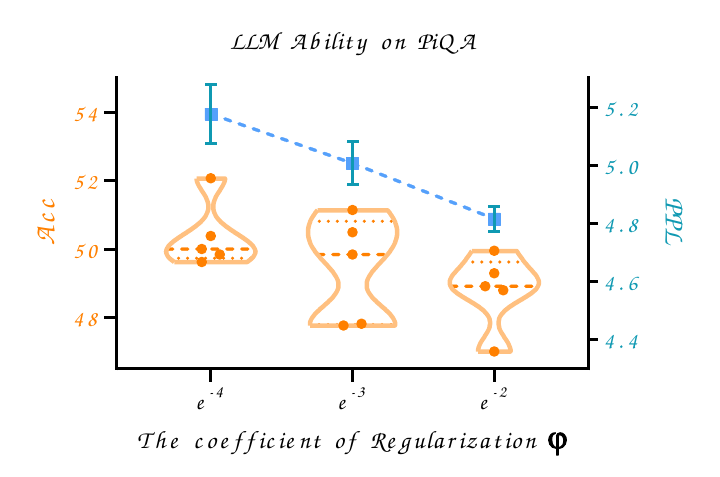}\label{fig:reg_conf_piqa}}
        \subfloat[MedMCQA]{\includegraphics[width=.33\textwidth]{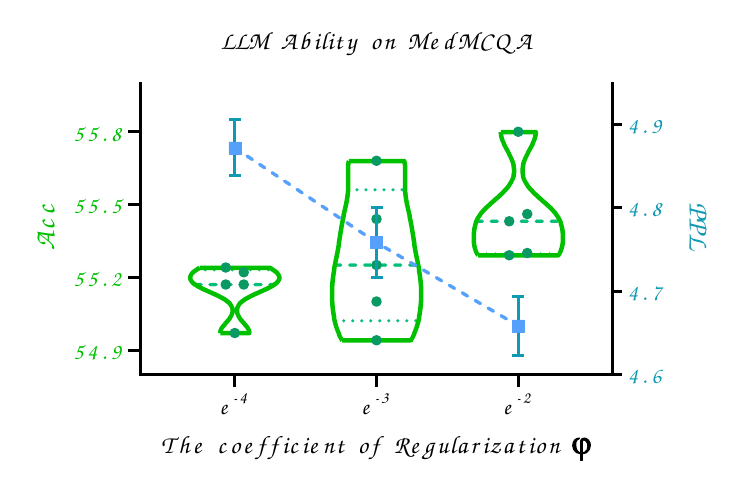}\label{fig:reg_conf_medmcqa}}
        \caption{The influence of regularization coefficient $\varphi$ on LLaMA-3 across datasets. (Acc$\uparrow$: Accuracy, PPL$\downarrow$: Perplexity.) }
        \label{fig:reg_conf_inf}
    \end{figure*}

% \begin{figure}[htbp]
%     \centering
%     % \begin{minipage}{0.49\linewidth}
%     %     \centering
%     %     \begin{subfig}
%     %         \centering
%     %         \includegraphics[width=\textwidth]{Figures/reg_conf/PPL.pdf}
%     %         \caption{XXX}
%     %         \label{fig:sub1}
%     %     \end{subfig}
%     %     % \includegraphics[width=\textwidth]{Figures/reg_conf/PPL.pdf}
%     %     % \vspace{0.1cm}
%     %     % \caption{General Ability}
%     %     % % {\small \hspace{0.5cm}General}  % 使用 \hspace 调整标题水平位置，并设置字体为小号
%     %     % \label{fig:image4}
%     % \end{minipage}
%     \begin{minipage}{0.49\linewidth}
%         \centering
%         \includegraphics[width=\textwidth]{Figures/reg_conf/Acc_PiQA.pdf}
%         \vspace{0.1cm} % 调整图片与标题之间的垂直间距
%         {\small \hspace{0.5cm}PiQA}  % 使用 \hspace 调整标题水平位置，并设置字体为小号
%         \label{fig:image1}
%     \end{minipage}%
%     \\
%     \begin{minipage}{0.49\linewidth}
%         \centering
%         \includegraphics[width=\textwidth]{Figures/reg_conf/Acc_SciQ.pdf}
%         \vspace{0.1cm}
%         {\small \hspace{0.5cm}SciQ}  % 使用 \hspace 调整标题水平位置，并设置字体为小号
%         \label{fig:image2}
%     \end{minipage}%
%     \begin{minipage}{0.49\linewidth}
%         \centering
%         \includegraphics[width=\textwidth]{Figures/reg_conf/Acc_MedMCQA.pdf}
%         \vspace{0.1cm}
%         {\small \hspace{0.5cm}MedMCQA}  % 使用 \hspace 调整标题水平位置，并设置字体为小号
%         \label{fig:image4}
%     \end{minipage}
%     % 
%     \caption{Regularization coefficient influence on LLaMA-3.}
%     \label{fig:reg_conf}
% \end{figure}

We compare our HLoRA method with the previous SOTA method, EWCLoRA, from two aspects: the time required for importance calculation and the storage memory needed. As shown in \cref{fig:Time Complexity}, our method is nearly \textbf{20 times faster} and requires only \textbf{10\%$\sim$15\% of the storage memory} compared to EWCLoRA, demonstrating the practicality of ours.

\textbf{Time Complexity:}
The experiments were conducted on an A800 GPU to evaluate the time complexity of our method in comparison with EWCLoRA. For EWCLoRA, the Fisher matrix computation followed the approach described in the original paper, using 20,000 randomly sampled data points from the Pile dataset with a maximum batch size of 8.
In contrast, for our method, the time measurement was based on 5 training epochs, a setting determined through empirical evaluation to achieve optimal performance.
% 实验结果显示，对于GPTJ-6B和LLaMA-3-3B，EWCLoRA分别需要27.17小时和25.97小时来计算重要性矩阵，而我们的HLoRA方法只要1.15和1.19小时。
The experimental results show that for GPT-J-6B and LLaMA-3-3B, EWCLoRA requires \textbf{27.17} and \textbf{25.97} hours, respectively, to compute the importance matrix, while our HLoRA method only takes \textbf{1.15} and \textbf{1.19} hours.
% The results demonstrate that our approach achieves a significant improvement in computational efficiency, requiring approximately \textbf{20 times less time} to calculate the importance matrix compared to EWCLoRA.

\textbf{Storage Memory:}
EWCLoRA necessitates the computation and storage of the Fisher matrix based on the Pile dataset before calculating the parameter importance.
According to the original paper, the Fisher matrix for GPT-J-6B occupies approximately \textbf{22.65} GB of memory. Similarly, for LLaMA-3-3B, the Fisher matrix occupies \textbf{11.97} GB of memory, calculated based on the Fisher computation method described in the original work.
In contrast, the storage memory of our method is only \textbf{3.5} GB and \textbf{1.3} GB, offering a significant advantage in terms of memory efficiency.  
This demonstrates that EWCLoRA incurs substantial storage overhead, whereas our method avoids such requirements, providing a more space-efficient solution.
    % EWC: GPTJ 22.65G; 11.96G
    % Ours:GPTJ 3.5G; LLaMA 1.3GF

\subsection{Regularization Coefficient Analysis}
\cref{fig:reg_conf_inf} demonstrate the effect of the regularization coefficient $\varphi$ in \cref{eq:total_loss} on PPL and accuracy across three tasks. As $\varphi$ increases, PPL gradually decreases, indicating a stronger emphasis on preserving general ability. Higher values of $\varphi$ correspond to better general ability retention. However, as shown in \cref{fig:reg_conf_piqa}, increasing $\varphi$ negatively impacts the average accuracy on PiQA.
Thus, $e^{-3}$ is selected as the optimal value for the regularization coefficient to balance task performance and general ability (lower PPL).

% \vspace{-0.3cm}
% \begin{figure}[htbp]
%     \centering
%     \begin{minipage}{0.9\linewidth}
%         \centering
%         \includegraphics[width=\textwidth]{Figures/imp_heatmap/llama_piqa.pdf}
%         \vspace{0.1cm} % 调整图片与标题之间的垂直间距
%         {\small \hspace{0.5cm}LLaMA-3}  % 使用 \hspace 调整标题水平位置，并设置字体为小号
%         \label{fig:image1}
%     \end{minipage}%
%     \\
%     \begin{minipage}{0.9\linewidth}
%         \centering
%         \includegraphics[width=\textwidth]{Figures/imp_heatmap/gptj_medmcqa.pdf}
%         \vspace{0.1cm}
%         {\small \hspace{0.5cm}GPT-J}  % 使用 \hspace 调整标题水平位置，并设置字体为小号
%         \label{fig:image4}
%     \end{minipage}
%     \caption{Log-scaled heatmap of L2 norms of parameter importance $\mathbf{\Omega}_l^\nu$ for q\_proj and v\_proj after LoRA fine-tuning on $\nu$ task across layers.}
%     \label{fig:importance_heatmap}
% \end{figure}

\begin{figure}[htb]
    \centering
    \subfloat[LLaMA-3]{
        \includegraphics[width=.9\linewidth]{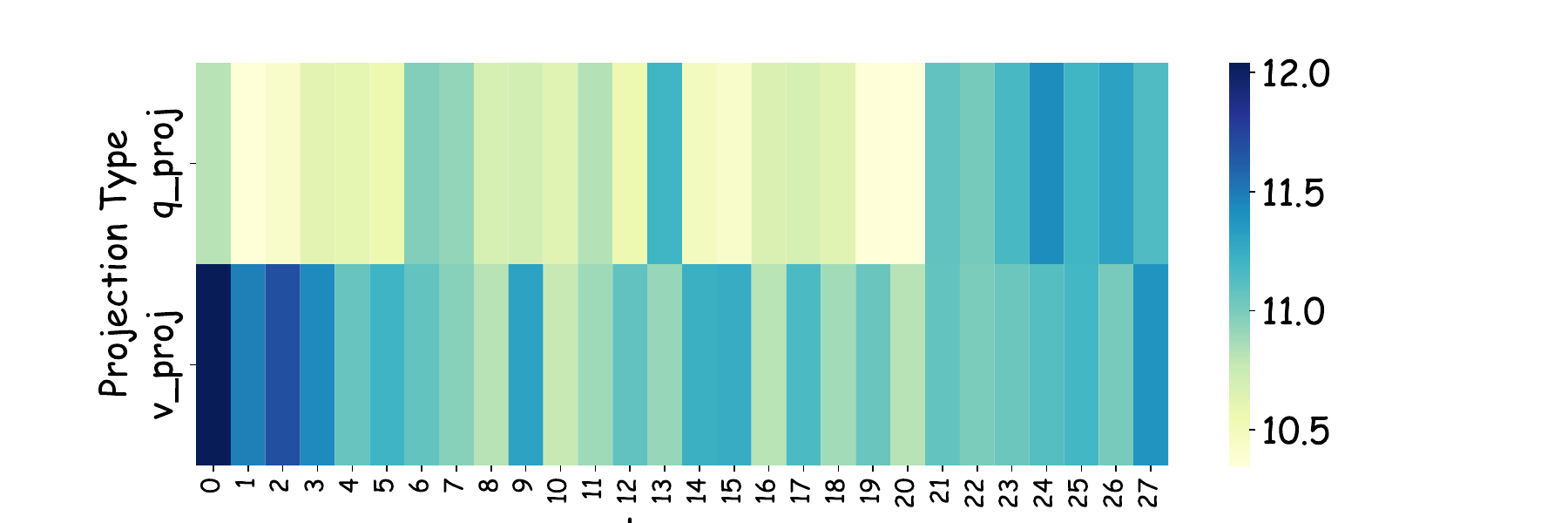}
        \label{fig:importance_heatmap_llama}
        }
    \\
    \subfloat[GPT-J]{
        \includegraphics[width=.9\linewidth]{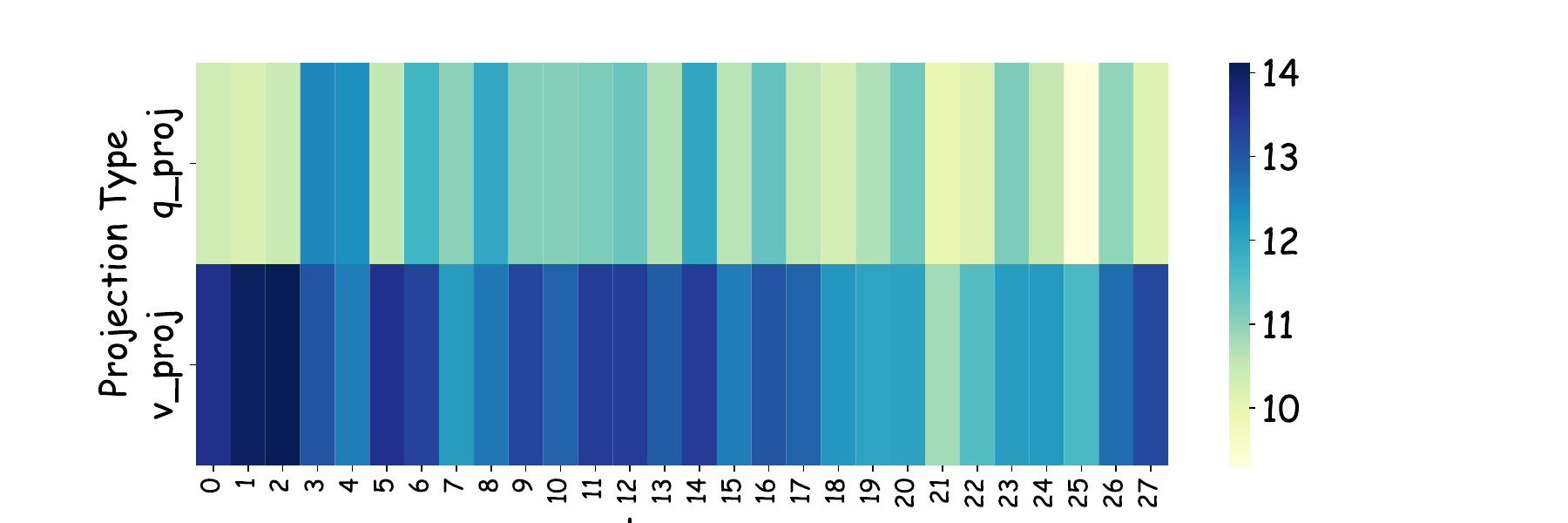}
        \label{fig:importance_heatmap_gptj}
        }
    \caption{Log-scaled heatmap of L2 norms of parameter importance $\mathbf{\Omega}_l^\nu$ for q\_proj and v\_proj after LoRA fine-tuning on $\nu$ task across layers.}
    \label{fig:importance_heatmap}
\end{figure}
% \vspace{-0.5cm}

\subsection{Parameters Importance Visualization}

    % "Log-Scaled Heatmap of L2 Norms for q\_proj and v\_proj During LoRA Fine-Tuning Across Layers"
    \cref{fig:importance_heatmap} highlights the importance in \cref{eq:importance} of q\_proj and v\_proj layers for general capabilities during the LoRA fine-tuning process. The heatmap illustrates that the v\_proj layers, particularly in the first four and the last layer, are crucial for preserving general knowledge. In contrast, the importance of the q\_proj layers is relatively weaker across the model. The L2 norms have been log-transformed to facilitate the comparison of the relative significance of these parameters across layers.

% Table generated by Excel2LaTeX from sheet 'Ablation'
% \begin{table*}[htb]
%     \centering
    
%     \resizebox{\linewidth}{!}{
%     \begin{tabular}{lcccccccccccc}
%     \toprule
%           & \multicolumn{6}{c}{\textbf{GPTJ}}            & \multicolumn{6}{c}{\textbf{LLaMA-3}} \\
%     \midrule
%     \multirow{2}[2]{*}{} & \multicolumn{2}{c}{SciQ} & \multicolumn{2}{c}{PiQA} & \multicolumn{2}{c}{Medmcqa} & \multicolumn{2}{c}{SciQ} & \multicolumn{2}{c}{PiQA} & \multicolumn{2}{c}{Medmcqa} \\
%           & G@P↓  & D@A↑  & G@P↓  & D@A↑  & G@P↓  & D@A↑  & G@P↓  & D@A↑  & G@P↓  & D@A↑  & G@P↓  & D@A↑ \\
%     \midrule
%     Ours  &       &       &       &       & 3.42  & 36.70  &       &       &       &       &       &  \\
%     w/o layer &       &       &       &       &       &       &       &       &       &       &       &  \\
%     w/o layer+element &       &       &       &       &       &       &       &       &       &       &       &  \\
%     \bottomrule
%     \end{tabular}%
%     }
%     \caption{Ablation Experiments.}
%     \label{tab:ablation}%
% \end{table*}%

% Table generated by Excel2LaTeX from sheet 'Ablation'
\begin{table}[htbp]
    \centering
    \caption{Ablation experiments. (layer: layer-wise weighted regularization, element: element-wise regularization.) }
    \resizebox{\linewidth}{!}{
    \begin{tabular}{lcccccc}
    \toprule
    \multirow{2}[2]{*}{} & \multicolumn{2}{c}{\textbf{SciQ}} & \multicolumn{2}{c}{\textbf{PiQA}} & \multicolumn{2}{c}{\textbf{MedMCQA}} \\
          & PPL↓  & Acc↑  & PPL↓  & Acc↑  & PPL↓  & Acc↑ \\
    \midrule
    \multicolumn{1}{c}{} & \multicolumn{6}{c}{\textbf{LLaMA-3}} \\
    \midrule
    Ours  & \textbf{4.64 } & \textbf{97.10 } & \textbf{4.90 } & \textbf{51.14 } & \textbf{4.64 } & \textbf{55.80 } \\
    - layer & 4.75  & 96.80  & 4.96  & 49.70  & 4.74  & 54.41  \\
    - layer, element & 5.31  & 96.10  & 5.58  & 46.91  & 5.15  & 53.12  \\
    \midrule
          & \multicolumn{6}{c}{\textbf{GPT-J}} \\
    \midrule
    Ours  &  \textbf{3.35}   & \textbf{96.80}  & \textbf{3.40}   & \textbf{50.49}  & \textbf{3.34}    & \textbf{36.10}    \\
    - layer                                        & 3.36            & 96.30           & 3.41            & 49.95           & 3.35             & 35.62             \\
    - layer, element                               & 3.39            & 96.20           & 3.52            & 49.51           & 3.37             & 33.66 \\
    \bottomrule
    \end{tabular}%
    }
  \label{tab:ablation}%
\end{table}%

\subsection{Ablation Study}
As shown in \cref{tab:ablation}, to investigate the role of different components in our proposed HLoRA, we conduct ablation studies by selectively removing certain structures and observing the resulting impact. Specifically, we exclude two sets of components: (1) \textit{layer}: eliminating the differentiation of importance among layers, and (2) \textit{layer, element}: 
removing both layer-wise and element-wise importance, i.e., training the $\nu$ task first and then training the $\mu$ task without imposing any regularization constraints throughout the process. 
Upon removing the two components, the performance of methods based on two backbone LLMs declines across three datasets, thereby highlighting the effectiveness of the layer-wise and element-wise importance introduced.

% The slowdown relative to HLoRA after disabling above components are shown in . We observe that layerwise importance 

\section{Conclusion}

This paper addresses the critical issue of catastrophic forgetting in large language models (LLMs) during domain-specific fine-tuning. We propose a novel fine-tuning framework that preserves general capabilities while enabling efficient adaptation to new domains, minimizing knowledge loss in tasks outside the fine-tuned domain.
Additionally, we introduce a layer-wise coefficient to adjust the balance between regularization loss and cross-entropy loss dynamically. This adjustment accounts for the varying contributions of different layers to both generalization and domain-specific learning. Extensive experiments in scientific, physical, and medical tasks show that our framework effectively mitigates catastrophic forgetting while maintaining performance in domain-specific tasks.

\clearpage

\section*{Impact Statements}
This paper presents work whose goal is to advance the field of Machine Learning. There are many potential societal consequences of our work, none which we feel must be specifically highlighted here.

\bibliography{main}
\bibliographystyle{icml2024}

%%%%%%%%%%%%%%%%%%%%%%%%%%%%%%%%%%%%%%%%%%%%%%%%%%%%%%%%%%%%%%%%%%%%%%%%%%%%%%%
%%%%%%%%%%%%%%%%%%%%%%%%%%%%%%%%%%%%%%%%%%%%%%%%%%%%%%%%%%%%%%%%%%%%%%%%%%%%%%%
% APPENDIX
%%%%%%%%%%%%%%%%%%%%%%%%%%%%%%%%%%%%%%%%%%%%%%%%%%%%%%%%%%%%%%%%%%%%%%%%%%%%%%%
%%%%%%%%%%%%%%%%%%%%%%%%%%%%%%%%%%%%%%%%%%%%%%%%%%%%%%%%%%%%%%%%%%%%%%%%%%%%%%%
\newpage
\appendix

% \section{Case Study}
% \input{Tables/casestudy}

% \onecolumn

% You can have as much text here as you want. The main body must be at most $8$ pages long.
% For the final version, one more page can be added.
% If you want, you can use an appendix like this one.  

% The $\mathtt{\backslash onecolumn}$ command above can be kept in place if you prefer a one-column appendix, or can be removed if you prefer a two-column appendix.  Apart from this possible change, the style (font size, spacing, margins, page numbering, etc.) should be kept the same as the main body.
% %%%%%%%%%%%%%%%%%%%%%%%%%%%%%%%%%%%%%%%%%%%%%%%%%%%%%%%%%%%%%%%%%%%%%%%%%%%%%%%
% %%%%%%%%%%%%%%%%%%%%%%%%%%%%%%%%%%%%%%%%%%%%%%%%%%%%%%%%%%%%%%%%%%%%%%%%%%%%%%%

\end{document}